\documentclass{article}
\usepackage{ijcai24}

\usepackage{times}
\usepackage{soul}
\usepackage{url}
\usepackage[hidelinks]{hyperref}
\usepackage[utf8]{inputenc}
\usepackage[small]{caption}
\usepackage{graphicx}
\usepackage{amsmath}
\usepackage{amsthm}
\usepackage{booktabs}
\usepackage[switch]{lineno}

\usepackage{amsfonts}
\usepackage{multirow}
\usepackage{enumitem}
\usepackage{subfigure}
\usepackage[ruled,linesnumbered]{algorithm2e}

\title{Imputation with Inter-Series Information from Prototypes \\ for Irregular Sampled Time Series }

\author{
  Zhihao Yu$^{*\dag}$
\and
Xu Chu$^{\ddag}$\and
Liantao Ma$^{* \text{§}}$\And
Yasha Wang$^{* \text{§}}$\And
Wenwu Zhu$^{\ddag}$\\
\affiliations
$^{*}$Key Laboratory of
High Confidence Software Technologies, Ministry of Education, Beijing, China \\
${^{\dag}}$School of Computer Science, Peking University, Beijing, China\\
$^{\ddag}$Department of Computer Science, Tsinghua University, Beijing, China\\
$^{\text{§}}$National Engineering Research Center of Software Engineering, Peking
University, Beijing, China\\
\emails
yuzhihao@stu.pku.edu.cn, chu\_xu@mail.tsinghua.edu.cn, \\malt@pku.edu.cn, wangyasha@pku.edu.cn, wwzhu@tsinghua.edu.cn
}

\begin{document}

\maketitle

\begin{abstract}
    Irregularly sampled time series are ubiquitous, presenting significant challenges for analysis due to missing values. Despite existing methods address imputation, they predominantly focus on leveraging intra-series information, neglecting the potential benefits that inter-series information could provide, such as reducing uncertainty and memorization effect. To bridge this gap, we propose PRIME, a \underline{P}rototype \underline{R}ecurrent \underline{I}mputation \underline{M}od\underline{E}l, which integrates both intra-series and inter-series information for imputing missing values in irregularly sampled time series. Our framework comprises a prototype memory module for learning inter-series information, a bidirectional gated recurrent unit utilizing prototype information for imputation, and an attentive prototypical refinement module for adjusting imputations. We conducted extensive experiments on three datasets, and the results underscore PRIME's superiority over the state-of-the-art models by up to 26\% relative improvement on mean square error. 
    Our code is available at \url{https://github.com/yzhHoward/PRIME}.
\end{abstract}

\section{Introduction}
Irregularly sampled time series are prevalent in various fields, such as healthcare \cite{goldberger2000physiobank} and meteorology \cite{schulz1997spectrum}. Taking Electronic Health Record (EHR) laboratory test data as an example, patients may undergo different tests during multiple visits, which leads to the absence of certain features during some visits. The irregularly sampled nature of this data brings missingness and impedes further analysis of these series, as most methods assume fully observed data \cite{berglund2015bidirectional,shukla2021multi}. Therefore, accurate imputation of irregularly sampled series is essential, serving as the cornerstone for subsequent analytical tasks.

In recent years, several methods \cite{cao2018brits,luo2019e2gan,shukla2018interpolation} have been proposed to impute irregularly sampled time series, utilizing techniques like time attention \cite{shukla2022heteroscedastic} or probabilistic sequence imputation \cite{fortuin2020gp}. However, these methods predominantly focus on intra-series information, overlooking inter-series relationships. In reality, there may be similarities between different series that can be leveraged to enhance imputation and reduce uncertainty. In particular, when imputing a series with a high missing rate, information from similar series can be valuable. Although seemingly straightforward, applying this intuition will face the following challenges:

\begin{itemize}
    \item \textbf{C1: It is difficult to identifying similar sequences.} Retrieving similar series from a dataset can be immensely time-consuming. If we directly retrieve raw sequences, it becomes challenging to identify similar sequences in smaller datasets. An alternative choice is to retrieve similar latent representations. Although the introduction of vector databases \cite{johnson2019billion} could accelerate the retrieval process, updating the database of latent representations during training still incur unsustainable time and storage costs.
    \item \textbf{C2: Information from similar sequences can be unreliable.} On the one hand, when there are plenty of missingness in retrieved similar sequences, information from them may be insufficient. On the other hand, the memorization effect \footnote{
        The memorization effect refers to the phenomenon where deep models firstly learn simple patterns and gradually memorize noise.} of deep models can lead to the model learning noise in the data \cite{arpit2017closer}. This problem can become more severe while using retrieving sequences for imputation. Since the encoding of similar series contains noise, noises can accumulate further during imputation using inter-series information.
\end{itemize}

To address these challenges, we propose capturing the pattern changes of sequences using prototypes and utilizing them as inter-series information to assist in imputing missing values in irregularly sampled series. Specifically, we design a Prototype Gated Recurrent Unit (P-GRU) for imputing, along with an Prototype Refinement module to correct the imputation. Notably, \textit{we use prototypes instead of retrieving similar series as external information}, which resolves the retrieval issue (C1). Besides, \textit{we capture the varying patterns from the imputed series in the context of learning the prototypes} and mitigate the impact of the memorization effect through innovative prototype losses (C2).

In summary, our model contributes in the following ways:
\begin{itemize}
    \item To the best of our knowledge, this work is the first to leverage prototypes to learn inter-series information for imputing missing values in irregularly sampled series.
    \item We introduce a Prototype Recurrent Imputation Model to capture inter-series information and refine prototypes for imputation.
    \item Extensive experiments on three datasets validate the effectiveness of the proposed method. Specifically, our experiments confirm that prototypes effectively alleviate the impact of the memorization effect.
\end{itemize}

\section{Related Work}

\subsection{Time Series Imputation}
Time series imputation is a classic problem in machine learning \cite{little2002single}. 
The naive solutions would be imputed with the mean value of all observed points or the front value from the last observation to impute missing values. However, the easy approaches could bring significant errors. Various methods have been developed to solve this problem. 
Li et al. \shortcite{li2009dynammo} find latent variables by fitting a Kalman Filter \cite{welch1995introduction} using the Expectation Maximization algorithm. 
Zhao et al. \shortcite{zhao2020missing} propose a semi-parametric algorithm that mixes data as a Gaussian copula to impute missing values. 
Yu et al. \shortcite{yu2016temporal} utilize matrix factorization with temporal regularization to impute the missing values in multivariate time series. 
For exploiting information from similar sequences, Rahman et al. \shortcite{rahman2014imputation} use the K-nearest-neighbors algorithm to find similar series in the dataset and adopt the mean value from neighbors to impute missing values. 
Wellenzohn et al. \shortcite{wellenzohn2017continuous} identify similar parts of the given series in history by Pearson's correlation coefficient and apply the mean value of matched blocks for imputation.

Inspired by the recent success of deep neural networks, many models based on RNNs or Transformers are implemented \cite{berglund2015bidirectional,kim2017recurrent,yoon2017multi,shukla2018interpolation,zerveas2021transformer,shukla2022heteroscedastic,du2022saits}. Che et al. \shortcite{che2018recurrent} propose GRU-D to incorporate mask and time interval information and mixes the last observed value and global mean value with a decay factor to fill missing values. 
A closely related work is BRITS \cite{cao2018brits}, which interpolates missing values through temporal decay and feature correlation in a bidirectional manner.

Generative methods \cite{ma2020adversarial,liu2019naomi,fortuin2020gp,luo2019e2gan,shukla2021multi} also play a role in time series imputation. 
Fortuin et al. \cite{fortuin2020gp} introduce a variational auto-encoder (VAE) architecture with a Gaussian process prior to embed the data into a smoother representation. 
Rubanova et al. \shortcite{rubanova2019latent} generalize RNNs to have continuous-time hidden dynamics defined by ordinary differential equations (ODE) inspired by Neural ODE \cite{chen2018neural} and perform interpolation tasks. 
Following the success generative adversarial network (GAN) \cite{goodfellow2020generative} in computer vision, 
Miao et al. \shortcite{miao2021generative} present a semi-supervised GAN based on the recurrent structure from BRITS \cite{cao2018brits} to impute with a fraction of labels. 
To model periodic and aperiodic information while encoding information about the sparsity of observation in the perspective of time, Shukla et al. \shortcite{shukla2022heteroscedastic} propose HeTVAE, which provides uncertainty of variables in interpolations.


\subsection{Prototype Learning}
Prototype learning is an empirical method in zero-shot and few-shot image classification. Each prototype belongs to a class and represents a high-level feature in this scenario. The model can efficiently classify a given image by the similarity between the image and prototypes 
\cite{snell2017prototypical,chen2018looks,doras2020prototypical}. 
Following the image classification methods, Zhang et al. \shortcite{zhang2020tapnet} employ prototypes on time series classification. Nevertheless, prototype learning has not been investigated in imputation of irregular sampled time series.

Most imputation works focus on intra-series information such as temporal information and feature interactions. 
In this work, prototype learning is introduced for the first time in time series imputation to leverage information from other series. Besides, we adaptively combine the information from prototypes with series, which solves the problem of individualized requirements for prototypes from different series.

\section{Preliminary}
Formally, the irregular sampled multivariate time series data $x$ can be defined as a series of $N$ features with $T$ observations, then the record of variable $n$ observed at time $t$ can be represented by $x_{t, n}$, where $1 \le t \le T$ and $1 \le n \le N$. For different series, the lengths $T$ can be different. In particular, the observation of $x_{t, n}$ is probably missing. To represent missing values, we introduce the mask $m_{t, n}$ to encode the missing state of $x_{t, n}$. Specifically, $m_{t, n} = 0$ indicates that the observation of the variable $n$ is missing at time step $t$ and $m_{t, n} = 1$ indicates that the variable $n$ is observed. Similar to Che et al. \shortcite{che2018recurrent}, we denote $\delta_{t, n}$ as the time gap between the last observation to the current timestamp $s_{t, n}$ of the feature $n$, where

\begin{equation}
\delta_{t, n}=
\begin{cases}
s_{t, n}-s_{t-1, n}+\delta_{t-1, n}& \text{if $t > 1, m_{t, n} = 0$} \\
s_{t, n}-s_{t-1, n}& \text{if $t > 1, m_{t, n} = 1$} \\
0& \text{if $t = 1$}
\end{cases}
\end{equation}

For example, giving timestamp $[0, 2, 3, 6, 10]$ of a series and missing mask $[1, 0, 0, 1, 1]$, the $\delta_t$ is $[0, 2, 3, 6, 4]$.

The time series imputation problem studied in this paper aims to find an appropriate value for each missing value and make the imputation $\hat{x}$ as close to $x$ as possible.


\section{Prototype Recurrent Imputation Model}
\begin{figure*}[h]
\centering
\includegraphics[width=1.7\columnwidth]{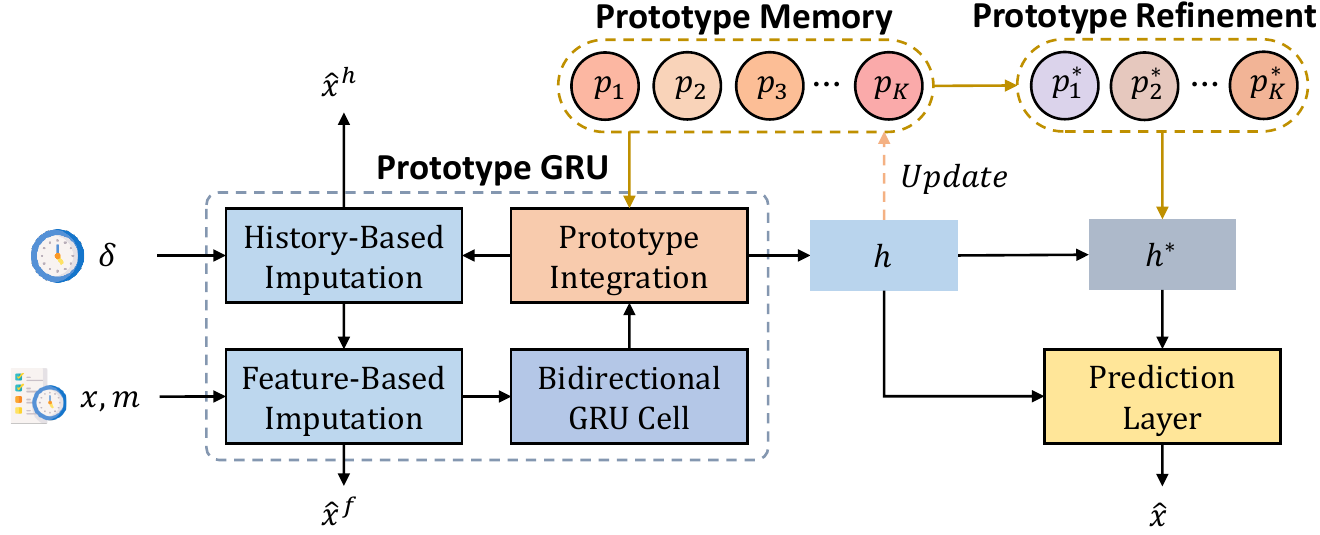}
\caption{The framework of the PRIME. Prototypes serve as inter-series information to assist in the imputation of time series in the bidirectional P-GRU and Prototype Refinement. The prototypes are updated with the representations from the time series}
\label{framework}
\end{figure*}

In this section, we present the proposed Prototype Recurrent Imputation Model (PRIME). The framework of PRIME is illustrated in Figure \ref{framework}. PRIME includes the following three detailed components:

\begin{itemize}[leftmargin=*]
\item The \textit{Prototype Memory} learns and stores prototypes. The prototypes are updated with latent representation of series by novel prototype losses.
\item The bidirectional \textit{Prototype Gated Recurrent Unit} (P-GRU) imputes missing values by flexible merging information from prototypes, historical patterns and feature interaction. 
\item The \textit{Prototype Refinement} localizes the prototypes by the imputed series from P-GRU and adjusts the imputation from a comprehensive view. 
\end{itemize}


\subsection{Prototype Memory}
We begin by discussing the prototype memory for learning inter-series information. To acquire series-level prototypes, a straightforward solution involves directly extracting information from the original sequences. However, when dealing with small datasets or datasets with substantial missingness, the information within the raw sequences is insufficient. Therefore, we leverage the latent representations of each time step embedded through bidirectional P-GRU steps to learn prototypes. This approach offers two advantages: firstly, P-GRU fills in the gaps in the sequence, ensuring reliable information; secondly, the representation at each time step consists of both forward and backward information, encompassing the entirety of the sequence information.

Formally, giving representations of series $\mathcal{S} \in \mathbb{R}^{M \times d}$, where $M$ is the number of representations equaling to the total time step in the dataset and $d$ is the dimension of hidden size, our model learns prototypes $\mathcal{P} = \{p_j\}^K_{j=1}$ from $\mathcal{S}$ where $p_j \in \mathbb{R}^{d}$ and $K$ is the number of prototypes.
Assuming that there are various types of series in the dataset, when prototypes and representations constitute a clustering relationship, the varying modes represented by prototypes are more stable and reliable.
To compose the cluster structure, we initialize prototypes with cluster centroids of $\mathcal{S}$. While training, we design cluster losses to update prototypes as the distribution shift of representations and maintain the cluster structure:
\begin{equation}
\begin{split}
\mathcal{L}_{clu} &= \lambda_{\mathcal{S} \rightarrow \mathcal{P}} \mathcal{L}_{\mathcal{S} \rightarrow \mathcal{P}} + \lambda_{\mathcal{P} \rightarrow \mathcal{S}} \mathcal{L}_{\mathcal{P} \rightarrow \mathcal{S}},\\
\mathcal{L}_{\mathcal{S} \rightarrow \mathcal{P}} &= \sum^M_{i} \mathop{\mathrm{min}}\limits_{j} \Vert s_i - p_j \Vert_2, \\
\mathcal{L}_{\mathcal{P} \rightarrow \mathcal{S}} &= \sum^K_{j} \mathop{\mathrm{min}}\limits_{i} \Vert s_i - p_j \Vert_2,
\end{split}
\end{equation}
The cluster losses consist of two parts. The first part assigns each representation to its nearest prototype and minimizes the distance between them. The other loss allocates each prototype to its nearest representation while guaranteeing that one representation corresponds to at most one prototype and minimizes their distance. These losses are only employed to learn prototypes and the gradients are not back-propagated to the other layers.

During experiments, we observe that with the above two losses, some prototypes may become homogenized during training (i.e., some prototypes are very close in the latent space), which will lead to a decrease in the diversity.
Therefore, we introduce a separating term $\mathcal{L}_{sep}$ to penalize pairs of prototypes that are close to each other and increase their distance with a margin. The cluster losses and the separation term together constitute the prototype losses.
\begin{equation}
\begin{split}
\mathcal{L}_{sep} = \sum^K_j \sum^{K}_{j' \ne j} \mathrm{max}(0, margin - \Vert p_j - p_{j'} \Vert_2).
\end{split}
\end{equation}

\subsection{Prototype Gated Recurrent Unit}
Time series analysis approaches based on RNN have achieved early success \cite{choi2016retain}. However, traditional RNN assumes that data are not missing and are observed at the same interval.
To leverage the advances of RNN while utilizing information from the learned prototypes, we propose the P-GRU to interpolate the missingness in every time step and encode sequence information. P-GRU is able to capture both intra-series information and inter-series information flexibly.


Given irregular sampled time series, P-GRU firstly impute missing values with the historical information and feature correlations. 

For modeling the historical information of the observation interval $\delta_t$ and representation of the last step $h_{t-1}$, we utilize a time decay term $\Delta_t$ and compute decayed latent feature $h^{\Delta}_{t}$ inspired by Che et al. \shortcite{che2018recurrent}. For $t=0$, $h_{t-1}$ is initialized with zeros.
\begin{equation}
\Delta_t = \mathrm{exp}(-\mathrm{max}(0, W_{\Delta} \delta_t)),
\end{equation}
\begin{equation}
h^{\Delta}_{t} = \Delta_t h_{t-1}.
\end{equation}
Then we interpolate based on decayed historical information and prototypical information by
\begin{equation}
\hat{x}^h_t = m_t * x_t + (1 - m_t) (W_h * h_{t}^{\Delta}),
\end{equation}
where $W_{\Delta}$ and $W_h$ are learnable parameters, and $*$ is element-wise multiplication.

Then we introduce a parameter matrix $W_f$ to capture the correlation across variates and obtain imputations $\hat{x}^f_t$ based on the correlation.
\begin{equation}
\hat{x}^f_t = m_t * x_t + (1 - m_t) * (W_f \hat{x}^h_t).
\end{equation}
The diagonal values of $W_f$ are set to zeros to suppress the auto-correlation and impute via the other variates.

With the imputed value $\hat{x}^f_t$, we use a vanilla GRUCell to embed information this step into latent space by
\begin{equation}
h^{intra}_t = \mathrm{GRUCell}((\hat{x}^f_t|m_t), h^{\Delta}_{t}),
\end{equation} where $(\cdot|\cdot)$ denotes the concatenation of two matrices.


After utilizing intra-series information for imputing missing values, we integrate this inter-series information to enhance the filling process in the subsequent time step. In the integration operation, P-GRU assigns the attention coefficient $\kappa \in \mathbb{R}^{K}$ of each prototype and integrate adaptively into the knowledge representation $h^{inter}_t$. The attention coefficient $\kappa$ enhances prototypes with strong correlations to $h^{intra}_t$ and suppresses the others.
\begin{equation}
\begin{split}
\kappa(h^{intra}_t, p) &= \frac{\mathrm{exp}(V_{\kappa} h^{intra}_t (W_{\kappa} p)^\top / \sqrt{d})}{\sum_j^K \mathrm{exp}(V_{\kappa} h^{intra}_t (W_{\kappa} p_j)^\top)}, \\
h^{inter}_t &= \kappa(h^{intra}_t, p) \cdot U_{\kappa} p.
\end{split}
\end{equation}
$U_{\kappa}$, $V_{\kappa}$, and $W_{\kappa} \in \mathbb{R}^{d \times d}$ are learnable transformation matrices. We employ the scaling factor $1/\sqrt{d}$ to normalize the dot product to counteract the growth of the magnitude.

Moreover, we combine local features $h^{intra}_t$ and prototype knowledge $h^{inter}_t$ into an integrated representation $h_t$. Instead of relying on manually tuned hyperparameters, we introduce a flexible fusion approach that extracts an appropriate amount of information from both the intra-series and inter-series perspectives based on the current time step's context. Thus, the model can adaptively determine the informational contributions from within and across series. To be more specific, a learnable weight parameter $\alpha$ is introduced to determine the proportion of the above-mentioned representations.

\begin{equation}
\begin{split}
\alpha &= \sigma(W_\alpha (h^{intra}_t|h^{inter}_t)), \\
h_{t} &= \alpha * h^{intra}_t + (1 - \alpha) * h^{inter}_t,
\end{split}
\end{equation}
where $\sigma$ represents the sigmoid function.

By integrating information from the prototypes, PRIME can enhance imputation performance and acquire improved representations. The learned representations are then employed to update the prototypes, establishing a positive feedback loop.

\subsection{Prototype Refinement}

While we have obtained history-based imputation $\hat{x}^h$ and feature-based imputation $\hat{x}^f$, the imputations are not entirely reliable as bidirectional P-GRU only provides stepwise fillings in two directions. In addition to employing prototypes in each step, we apply prototypes to revise and offer imputations from a holistic perspective. 

Due to instance variations, prototypes may not adequately capture fine-grained information of all series. Therefore, we adaptively refine the prototype $p$ to $p^*$ by interacting with the information within the given series. Refined prototypes, compared to the original ones, incorporate information specific to the given series and provide more helpful knowledge in subsequent interactions.



Concretely, with the prototypes $\mathcal{P} = \{p_1, p_2, ..., p_K\}$ and overall latent features $h = [h_1, h_2, ..., h_T]$ from bidirectional P-GRU, we generate refined prototypes $\mathcal{P}^* = \{p^*_1, p^*_2, ..., p^*_K\}$ for each series via:
\begin{equation}
\begin{split}
p^* &= \zeta(p, h) \cdot U_{\zeta} h, \\
\zeta(p, h) &= \frac{\mathrm{exp}(V_{\zeta} p (W_{\zeta} h)^\top / \sqrt{d})}{\sum_t^T \mathrm{exp}(V_{\zeta} p (W_{\zeta} h_t)^\top)}.
\end{split}
\end{equation}

Then we calculate the correlation $\xi(h, p^*)$ between the overall representation $h$ and refined prototypes $p^*$ similar to Equation 5 through
\begin{equation}
\xi(h, p^*) = \frac{\mathrm{exp}(V_{\xi} h (W_{\xi} p^*)^\top / \sqrt{d})}{\sum_j^K \mathrm{exp}(V_{\xi} h (W_{\xi} p^*_j)^\top)},
\end{equation}
and the fused representation $h^*$ can be calculated by
\begin{equation}
h^* = \xi(h, p^*) \cdot U_{\xi} p^*.
\end{equation}

With the fused representation, we perform the final imputation on missing values as $\hat{x}$: 
\begin{equation}
\hat{x} = \mathrm{MLP}((h|h^*)).
\end{equation}
MLP is a two-layer perceptron with GELU \cite{hendrycks2016gaussian} as the activation function to model non-linear information.

\subsection{Learning Objective}
We adopt mean square error as the main optimization object for imputation.
With the imputations $\hat{x}^h$ and $\hat{x}^f$ from forward and backward prototype GRU, the imputation losses are
\begin{equation}
\mathcal{L}_{final} = (x - \hat{x})^2,
\end{equation}
\begin{equation}
\mathcal{L}_{forward} = (x - \mathop{\hat{x}^h} \limits ^{\rightarrow})^2 + (x - \mathop{\hat{x}^f} \limits ^{\rightarrow})^2,
\end{equation}
\begin{equation}
\mathcal{L}_{backward} = (x - \mathop{\hat{x}^h} \limits ^{\leftarrow})^2 + (x - \mathop{\hat{x}^f} \limits ^{\leftarrow})^2.
\end{equation}
And the overall learning objective is 
\begin{equation}
\mathcal{L} = \mathcal{L}_{final} + \lambda\mathcal{L}_{forward} + \lambda\mathcal{L}_{backward} + \mathcal{L}_{clu} + \lambda_{sep}\mathcal{L}_{sep},
\end{equation}
where $\lambda$ and $\lambda_{sep}$ are hyper-parameters to control the loss weight of imputations from P-GRU and prototype separating term.

\section{Experiments}

In this section, we perform interpolating experiments on three real-world datasets and compare the proposed model with the latest state-of-the-art methods. Then we conduct analyses showing the effectiveness of our intuition in employing the proposed modules. 

\subsection{Datasets}
In this section, we conduct experiments on three real-world datasets, namely PhysioNet Challenge 2012 \cite{goldberger2000physiobank}, PhysioNet Challenge 2019 \cite{reyna2019early}, and Human Activity following previous works \cite{cao2018brits,miao2021generative,shukla2022heteroscedastic}. The statistics of these datasets are in Table \ref{dataset}. Details of these datasets can be found in the Appendix.

\begin{table}[h]
  \caption{Statistics of PhysioNet Challenge 2012 (C12), Physionet Challenge 2019 (C19) and Human Activity (Activity).}
  \centering
  \begin{tabular}{lccc}
    \toprule
     & C12 & C19 & Activity \\
    \midrule
    Sample & 11,988 & 40,335 & 6,554 \\
    Variables & 37 & 34 & 12 \\
    Total Length & 560,294 & 1,552,160 & 327,700 \\
    Maximum Length & 48 & 336 & 50 \\
    Minimum Length & 1 & 8 & 50 \\
    Observation Rate & 24.72\% & 19.87\% & 25.00\% \\
    \bottomrule
  \end{tabular}
  \label{dataset}
\end{table}

\subsection{Evaluation Settings}
We assess the mean square error (MSE) and mean absolute error (MAE) of imputations on these datasets. The datasets are into the training set, verification set, and test set according to the ratio of 8:1:1. The model achieves lowest MSE on the evaluation dataset will be used on the test set. We randomly mask 10\% of available observations for imputation. To evaluate precisely, we repeat every experiment 3 times with different random seeds and compute the standard variance. The generated masks for imputing are different on different seeds. 
The hidden size $d$ of all modules in PRIME is 128. We set $\lambda = 0.3$, $\lambda_{\mathcal{S} \rightarrow \mathcal{P}} = 1$, $\lambda_{\mathcal{P} \rightarrow \mathcal{S}} = 0.1$, $\lambda_{sep} = 0.1$, and $\displaystyle margin = 50 / \sqrt{K}$ for all the datasets. The number of prototypes $K$ on these datasets is 64. We employ a modified Jonker-Volgenant algorithm \cite{crouse2016implementing} to ensure one fragment corresponds to at most one prototype in $\mathcal{L}_{\mathcal{P} \rightarrow \mathcal{S}}$.

We compare with the following baselines:

\begin{itemize}[leftmargin=*]
    \item Mean: The missing values are simply replaced with the global mean of the corresponding variate.
    \item KNN \cite{fancyimpute}: K-nearest neighbor algorithm is applied to search the similar series for the given series and impute with the average of neighbors.
    \item MF \cite{fancyimpute}: Direct matrix factorizes the incomplete matrix into two low-rank matrices and imputes by matrix completion.
    \item GRU-D \cite{che2018recurrent}: GRU-D exploits representations of time intervals and interpolates with weighted combination of the last observation and the global mean.
    \item BRITS \cite{cao2018brits}: BRITS imputes by capturing time intervals and feature correlation with a bidirectional RNN structure.
    \item SSGAN \cite{miao2021generative}: SSGAN is a generative adversarial network based on bidirectional RNN.
    \item HeTVAE \cite{shukla2022heteroscedastic}: HeTVAE is a Transformer-based model to embed sparsity with a temporal VAE architecture to propagate uncertainty.
\end{itemize}

For a fair comparison, all the hyper-parameters are fine-tuned through a grid search strategy. 



\begin{table*}[h]
  \caption{Performance comparison of MSE and MAE with standard variance on the three datasets}
  \centering
  \resizebox{\linewidth}{!}{
  \begin{tabular}{llcccccc}
    \toprule
    & & \multicolumn{2}{c}{PhysioNet Challenge 2012} & \multicolumn{2}{c}{PhysioNet Challenge 2019} & \multicolumn{2}{c}{Human Activity}\\
    & Model & MSE & MAE & MSE & MAE & MSE & MAE\\
    \midrule
    & Mean & 0.954$\pm$0.010 & 0.702$\pm$0.001 & 0.976$\pm$0.012 & 0.741$\pm$0.002 & 0.997$\pm$0.020 & 0.810$\pm$0.009\\
    & KNN & 0.666$\pm$0.033 & 0.532$\pm$0.027 & 0.640$\pm$0.011 & 0.557$\pm$0.001 & 0.340$\pm$0.005 & 0.386$\pm$0.004\\
    & MF & 0.433$\pm$0.009 & 0.402$\pm$0.001 & 0.449$\pm$0.011 & 0.447$\pm$0.002 & 0.257$\pm$0.007 & 0.332$\pm$0.002\\
    Baseline & GRU-D & 0.632$\pm$0.044 & 0.494$\pm$0.018 & 0.682$\pm$0.055 & 0.574$\pm$0.031 & 0.569$\pm$0.041 & 0.484$\pm$0.037\\
    & BRITS & 0.374$\pm$0.006 & 0.317$\pm$0.016 & 0.346$\pm$0.015 & 0.347$\pm$0.001 & 0.289$\pm$0.005 & 0.335$\pm$0.001\\
    & SSGAN & 0.341$\pm$0.005 & 0.284$\pm$0.009 & 0.302$\pm$0.015 & 0.326$\pm$0.001 & 0.240$\pm$0.006 & 0.299$\pm$0.002\\
    & HeTVAE & 0.328$\pm$0.017 & 0.331$\pm$0.011 & 0.319$\pm$0.006 & 0.371$\pm$0.004 & 0.168$\pm$0.007 & 0.253$\pm$0.004\\
    \midrule
    & P-GRU$_{inter-}$ & 0.278$\pm$0.011 & 0.274$\pm$0.001 & 0.273$\pm$0.011 & 0.315$\pm$0.004 & 0.171$\pm$0.008 & 0.264$\pm$0.003\\
    & P-GRU & 0.269$\pm$0.015 & 0.273$\pm$0.002 & 0.263$\pm$0.006 & 0.312$\pm$0.001 & 0.162$\pm$0.010 & 0.255$\pm$0.005\\
    Proposed & PRIME$_{refine-}$ & 0.271$\pm$0.010 & 0.270$\pm$0.003 & 0.261$\pm$0.010 & 0.312$\pm$0.003 & 0.162$\pm$0.009 & 0.252$\pm$0.004\\
    & PRIME & \textbf{0.243$\pm$0.009} & \textbf{0.260$\pm$0.001} & \textbf{0.250$\pm$0.010} & \textbf{0.306$\pm$0.003} & \textbf{0.157$\pm$0.008} & \textbf{0.247$\pm$0.004}\\
    \bottomrule
  \end{tabular}
  }
  \label{result}
\end{table*}

\subsection{Results and Analysis}
Table \ref{result} shows the MSE and MAE with standard variance results on the three datasets. Non-parameter algorithms (mean imputation and KNN) fail to achieve high accuracy on these datasets. 
The poor performance of GRU-D suggests that the information from the last step, time interval, and the global mean value is not enough to interpolate. By leveraging bidirectional historical information and feature interaction, BRITS and SSGAN bring a large improvement. The temporal attention with sparsity encoding promotes the performance of HeTVAE. Though recent deep learning models achieve satisfactory results, PRIME demonstrates the benefits of prototypes to exploit inter-series information.

On the PhysioNet Challenge 2012 dataset, compared to the previous state-of-the-art model, HeTVAE, PRIME reduces MSE by more than 26.0\% relatively. On the PhysioNet Challenge 2019 dataset, compared to the best baseline SSGAN, our model improves by 17.3\% on MSE. The variation and missing patterns in the human activity dataset are relatively simple, so the baseline method achieves better imputation performance, with a relative advantage of 6.1\% for PRIME on MSE.
Besides, the standard variance on PRIME is smaller than the others, showing the stability of PRIME. 

\subsection{Ablation Study}
To explore the effectiveness of the proposed modules, we conduct the following ablation studies:
\begin{itemize}[leftmargin=*]
  \item P-GRU$_{inter-}$: We evaluate P-GRU without prototypes. In this way, the imputations are computed without any information inter-series.
  \item P-GRU: We assess the bidirectional P-GRU alone to evaluate the benefit of the inter-series information and final imputation by Equation 15. 
  \item PRIME$_{refine-}$: We remove the prototype refinement operation $\zeta(p, h)$ and compute $\xi(h, p)$ instead of $\xi(h, p^*)$ in the interaction between prototypes and the entire sequence to examine the effect of refined prototypes.
\end{itemize}

In Table \ref{result}, comparing P-GRU$_{p-}$ and P-GRU, there is an extensive performance decay without prototypes. Moreover, PRIME noticeably outperforms P-GRU and PRIME$_{refine-}$, demonstrating that the fine-grained revisions from refined prototypes work effectively on interpolating missing values. On the PhysioNet Challenge 2012 dataset, the original prototypes do not boost the performance of the attention module. This is because the coarse-grained information represented by the prototypes has been utilized in P-GRU, and adjusting the prototypes by the correlation between the given series and prototypes is fruitful.

\subsection{Further Analyses and Observations}

\subsubsection{Using prototypes can combat memorization effect}
To explore whether the prototypes mitigate noises in the dataset, we design an experiment about the influence of capturing reliable patterns in different training periods. This experiment can also verify the memorization effect that neural networks first learn simple patterns and then learn noises gradually in the training process \cite{arpit2017closer}. 

\begin{figure}[h]
    \centering
    \includegraphics[width=0.9\columnwidth]{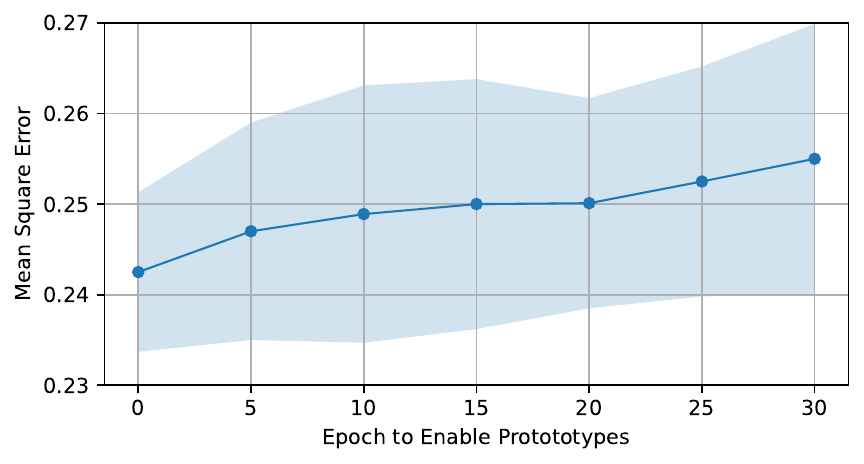}
    \caption{The results of different starting epochs to learn and utilize prototypes. The shadow represents the standard variance. The later the prototype is introduced, the worse the performance}
    \label{epoch}
\end{figure}

We start to integrate the information from prototypes and series information at epochs \{0, 5, 10, 15, 20, 25, 30\} instead of introducing prototypes at the beginning. The procedure of learning prototypes is performed throughout as usual. During the first several training epochs, we set $h_{t}=h^{intra}_{t}$ in P-GRU and $h^*$ to zeros vector to disable prototypes. In this way, the prototypes cannot be utilized until we enable them.

According to the results in Figure \ref{epoch}, the imputation performance decreases if prototypes are employed later, confirming that the model captures more noises in data as training. Since $\mathcal{L}_{clu}$ cannot update the model other than the prototype, the noise accumulates more and makes the prototype less reliable when it is adopted later. This experiment illustrates that prototypes denoise successfully, and the model learns minimal noise early in training.

\begin{figure}[h]
    \centering
    \includegraphics[width=0.9\columnwidth]{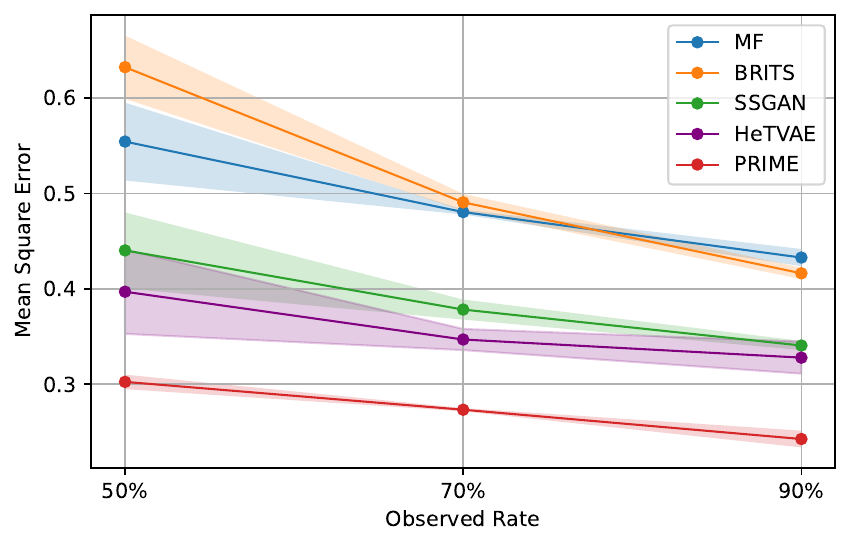}
    \caption{Results for different observation rates on the PhysioNet Challenge 2012 dataset}
    \label{missing}
\end{figure}

\begin{figure*}[h]
    \centering
    \subfigure {
    \includegraphics[width=0.65\columnwidth]{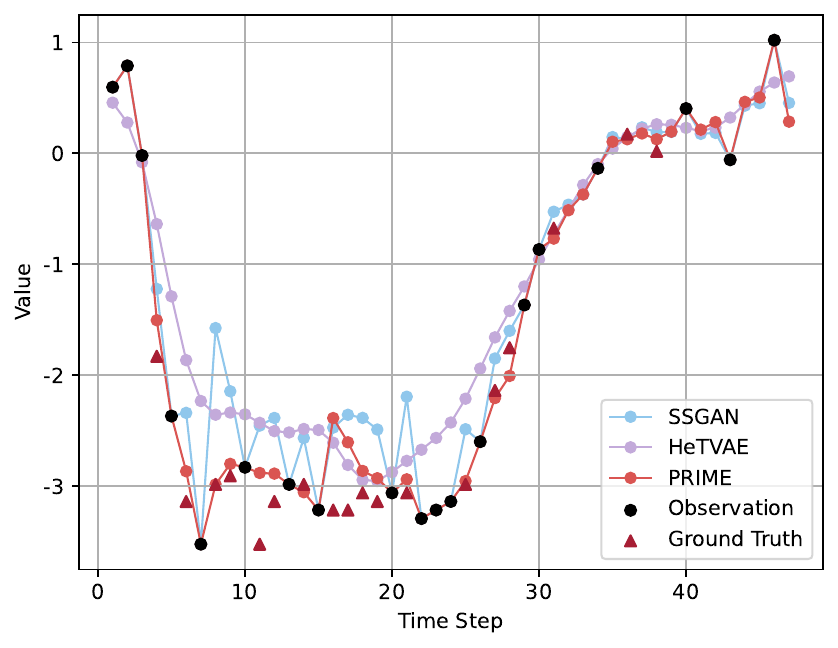}  
    }
    \subfigure {
    \includegraphics[width=0.65\columnwidth]{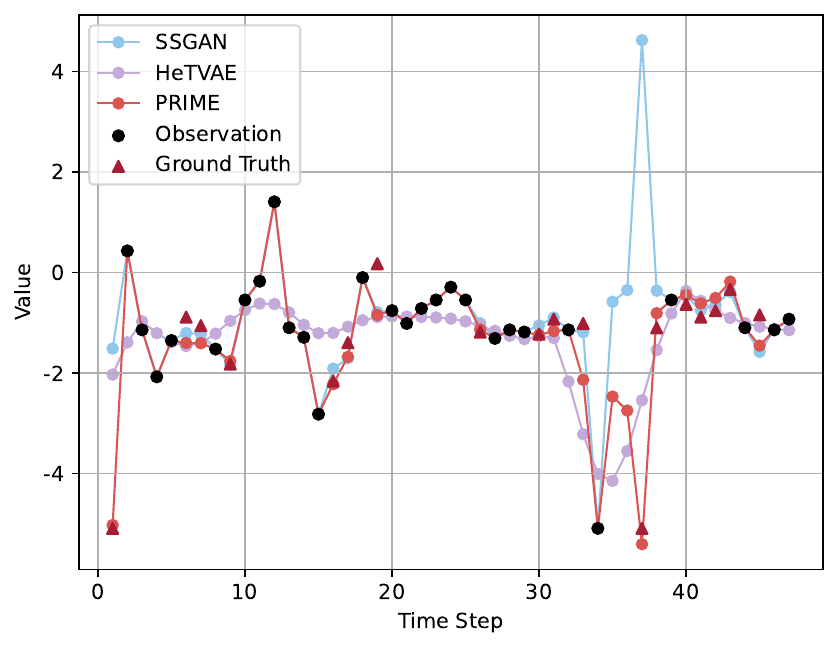}  
    }
    \subfigure {
    \includegraphics[width=0.65\columnwidth]{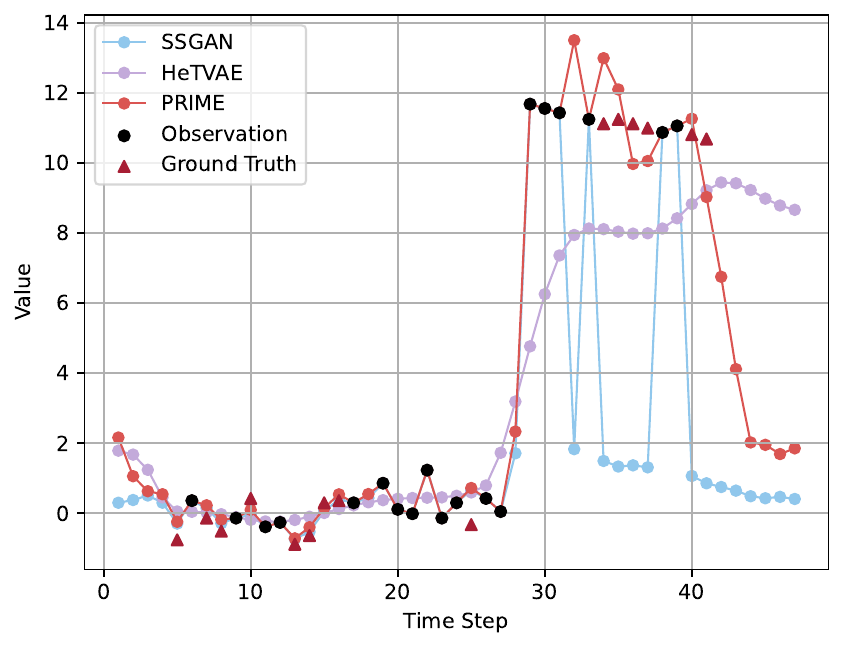}  
    }
    \caption{Examples of imputation results on the Challenge 2012 dataset with 50\% missing observations. For better observation, the values of SSGAN and PRIME are set to the existing observing data if available}
    \label{visualization}
\end{figure*}



\subsubsection{PRIME performs better on lower observation rate}
In order to evaluate the robustness of the proposed model under different missing rates, we conduct experiments by varying observation rates from \{50\%, 70\%, 90\%\} on Physionet Challenge 2012 dataset. 
The results are reported in Figure \ref{missing}.

PRIME greatly exceeds the baselines on these observation rates. Compared to the well-performing baselines, SSGAN and HeTVAE, PRIME brings more notable improvements with higher observation rates. Moreover, the standard variance of PRIME is smaller, indicating that our model's performance is robust though with more missingness and different missing position in different seeds.
These results illustrate that incorporating inter-series information can utilize the existing information more efficiently and estimate more precisely in high-missing situations. 

\subsubsection{PRIME is stable to the number of prototypes} 
\begin{figure}[h]
  \centering
  \includegraphics[width=0.9\columnwidth]{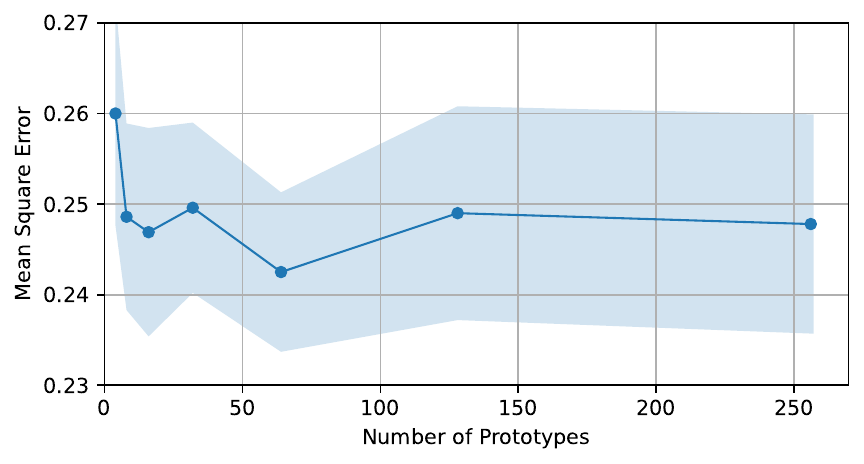}
  \caption{Results for different numbers of prototypes on the PhysioNet Challenge 2012 dataset}
  \label{prototype}
\end{figure}

To further explore the influence of the number of prototypes, we set $K = \{4, 8, 16, 32, 64, 128, 256\}$ and evaluate on the Physionet Challenge 2012 dataset. 
The results in Figure \ref{prototype} prove that our model is not sensitive to the number of prototypes from a considerable interval. 
However, too few prototypes ($K=4$) will lead to a degradation in performance since the inter-series information and diversity from prototypes represented by prototypes are insufficient.
The stability of PRIME is contributed by: (1)
The learnable weight $\alpha$ integrates inter-series and intra-series information adaptively in Equation 10. (2) The concatenate operation fuses representations $h$ and $h^*$ in Equation 14.

\subsubsection{All of the proposed prototype losses are effective}
The prototype losses $\mathcal{L}_{clu}$ and $\mathcal{L}_{sep}$ in PRIME promote the cluster structure between prototypes and the latent representations of time series.
To validate the efficiency of the proposed prototype losses, we conduct an ablation experiment. In this experiment, we try various combinations of them and the results are shown in Table \ref{loss}. By observing the results, we can see that removing any of the prototype losses results in performance degradation, which proves their effectiveness. The three loss functions play different roles in facilitating the prototypes and maintaining the clustering relationships between representations of series and prototypes.

\begin{table}[h]
  \caption{Different hyper-parameters of PRIME on PhysioNet Challenge 2012}
  \centering
  \begin{tabular}{ccccc}
    \toprule
    $\mathcal{L}_{\mathcal{S} \rightarrow \mathcal{P}}$ & $\mathcal{L}_{\mathcal{P} \rightarrow \mathcal{S}}$ &
    $\mathcal{L}_{sep}$ & MSE & MAE\\
    \midrule
    \checkmark & \checkmark & \checkmark & \textbf{0.243$\pm$0.009} & \textbf{0.260$\pm$0.001} \\
    $\times$ & \checkmark & \checkmark & 0.247$\pm$0.014 & 0.261$\pm$0.002 \\
    \checkmark & $\times$ & \checkmark & 0.247$\pm$0.010 & 0.261$\pm$0.002 \\
    \checkmark & \checkmark & $\times$ & 0.246$\pm$0.012 & 0.261$\pm$0.002 \\
    $\times$ & $\times$ & \checkmark & 0.247$\pm$0.011 & 0.260$\pm$0.001 \\
    $\times$ & $\times$ & $\times$ & 0.254$\pm$0.014 & 0.260$\pm$0.002 \\
    \bottomrule
  \end{tabular}
  \label{loss}
\end{table}

\subsubsection{Imputations from PRIME are efficacious qualitatively}
To verify the imputations qualitatively, we compare the imputation cases on the Challenge 2012 dataset with 50\% missing observations in Figure \ref{visualization}. The series are from different variables in different data. 
For better comparison, the values of SSGAN and PRIME are set to the existing observing data if available. 
As we can see, the interpolations from HeTVAE are smooth due to its heteroscedastic layer. However, this smoothing property leads to significant imputing errors when the trend of the series changes rapidly. Although SSGAN can fit rapid changes, it can be easily confused by the few observations and produce inaccurate interpolations. PRIME provides the closest imputations to the ground truth, which confirm the efficiency of alleviating inter-series information.


\section{Conclusion}
In this paper, we introduce PRIME, a prototype recurrent imputation model to capture and employ inter-series information in irregular sampled time series imputation. 
PRIME contains prototype memory, a bidirectional prototype gated recurrent unit, and prototype refinement. To learn representative and diverse prototypes, we develop novel clustering and separating losses that keep prototypes and series' representations to conform the clustering structure. The P-GRU can exploit both intra-series and inter-series information to impute missing values step by step. For the purpose of revising the imputations in a comprehensive view, we create prototype refinement for regulating the prototypes according to the given series and provide fine-grained guidance. 

Through extensive experiments on three datasets, we demonstrate that our model achieves superior performance over state-of-the-art methods by a significant improvement with high robustness. 
We sincerely hope the proposed method can provide valuable insights for downstream analysis of irregular sampled time series.

\clearpage

\bibliographystyle{named}
\bibliography{reference}

\end{document}